\documentclass[11pt,a4paper]{article}
\usepackage[draft]{hyperref}
\usepackage[hyperref]{emnlp2018}
\usepackage{times}
\usepackage{latexsym}
\usepackage{amsmath}
\usepackage{multirow}
\usepackage{url}

\aclfinalcopy 

\usepackage{graphicx} 
\usepackage{caption}
\usepackage{subcaption}
\usepackage{booktabs}
\usepackage{tabularx}
\usepackage{amsmath}
\usepackage{amssymb}
\usepackage[warn]{textcomp}
\usepackage{framed}
\usepackage{enumitem}
\usepackage{tikz}

\DeclareMathOperator{\relu}{ReLU}
\newcounter{tblexamplesrowcounter}
\newcommand\tblexamplesrownumber{\stepcounter{tblexamplesrowcounter}\arabic{tblexamplesrowcounter}}

\title{Lessons from Natural Language Inference \\ in the Clinical Domain}

\author{Alexey Romanov \\
  Department of Computer Science \\
  University of Massachusetts Lowell\thanks{\quad Work done during an internship at IBM Research}  \\
  Lowell, MA 01854\\ 
  {\tt aromanov@cs.uml.edu} \\\And
  Chaitanya Shivade \\
  IBM Almaden Research Center \\
  650 Harry Road \\
  San Jose, CA 95120 \\
  {\tt cshivade@us.ibm.com} \\}

\date{}

\AddToShipoutPicture{\put(555,350){\rotatebox{90}{\scalebox{1}{Extended version of the EMNLP 2018 paper.}}}}

\begin{document}
\maketitle
\begin{abstract}
  State of the art models using deep neural networks have become very good in learning an accurate mapping from inputs to outputs. However, they still lack generalization capabilities in conditions that differ from the ones encountered during training. This is even more challenging in specialized, and knowledge intensive domains, where training data is limited. To address this gap, we introduce MedNLI\footnote{\url{https://jgc128.github.io/mednli/}} -- a dataset annotated by doctors, performing a natural language inference task (NLI), grounded in the medical history of patients. We present strategies to: 1) leverage transfer learning using datasets from the open domain, (e.g. SNLI) and 2) incorporate domain knowledge from external data and lexical sources (e.g. medical terminologies). Our results demonstrate performance gains using both strategies.
\end{abstract}

\section{Introduction}
%
Natural language inference (NLI) is the task of determining whether a given \textit{hypothesis} can be inferred from a given \textit{premise}. This task, formerly known as recognizing textual entailment (RTE) \cite{dagan2006pascal} has long been a popular task among researchers. Moreover, contribution of datasets from past shared tasks \cite{dagan2009recognizing}, and recent research \cite{snli,multinli} have pushed the boundaries for this seemingly simple, but challenging problem.

The Stanford Natural Language Inference (SNLI) dataset~\cite{snli} is a large, high quality dataset and serves as a benchmark to evaluate NLI systems. However, it is restricted to a single text genre (Flickr image captions) and mostly consists of short and simple sentences. The MultiNLI corpus~\cite{multinli} which introduced NLI corpora from multiple genres (e.g. fiction, travel) was a welcome step towards addressing these limitations. MultiNLI offers diversity in linguistic phenomena, which makes it more challenging.

Patient data is guarded by careful access protection due to its sensitive content. Therefore, the common approach of using crowd sourcing platforms to get annotations is not possible in this domain. Moreover, labeling requires domain experts increasing the costs of annotation. Owing to these restrictions, the onset of community-driven NLP research facilitated by shared resources has been late in the clinical domain. Despite these barriers, publicly available datasets have been created through initiatives such as i2b2\footnote{http://www.i2b2.org/NLP} and CLEF.\footnote{http://www.clef-initiative.eu} However, most of these datasets have modest sizes, and they either target fundamental NLP problems (e.g. co-reference resolution) or information extraction tasks (e.g. named entity extraction). Currently, the clinical domain lacks large labeled datasets to train modern data-intensive models for end-to-end tasks such as NLI, question answering, or paraphrasing. 

Previous research relevant to the present topic, is the work on RTE in the biomedical domain: automatic construction of textual entailment datasets \cite{abacha2015semantic,abacha2016recognizing}, use of active learning on limited RTE data \cite{shivade2015textual,shivade2016addressing}, and enhancement of search results \cite{adler2012entailment} using TE models. These efforts were limited due to the above-mentioned constraints. Most importantly, none of datasets used in the above studies are publicly available.

To this end, we explore the problem of NLI in the clinical domain. Language inference in specialized domains such as medicine is extremely complex and remains unexplored by the machine learning community. Moreover, since this domain has a distinct sublanguage \cite{friedman2002two}, clinical text it also presents unique challenges (abbreviations, inconsistent punctuation, misspellings, etc.) that differentiate it from open-domain data~\cite{meystre2008Extracting}.

\begin{table*}[ht]
\footnotesize
\centering
\begin{tabularx}{\linewidth}{@{}lXXl@{}}
\toprule
\textbf{\#} & \textbf{Premise}                                                                                                                           & \textbf{Hypothesis}                                                   & \textbf{Label} \\ \midrule

\tblexamplesrownumber & ALT , AST , and lactate were elevated as noted above & patient has abnormal lfts & entailment \\
\tblexamplesrownumber & Chest x-ray showed mild congestive heart failure & The patient complains of cough & neutral \\
 \tblexamplesrownumber & During hospitalization , patient became progressively more dyspnic requiring BiPAP and then a NRB & The patient is on room air & contradiction \\
 \tblexamplesrownumber & She was not able to speak , but appeared to comprehend well & Patient had aphasia & entailment \\
 \tblexamplesrownumber & T1DM : x 7yrs , h/o DKA x 6 attributed to poor medication compliance , last A1c [ ** 3-23 ** ] : 13.3~\%~2 & The patient maintains strict glucose control & contradiction \\
 \tblexamplesrownumber & Had an ultimately negative esophagogastroduodenoscopy and colonoscopy & Patient has no pain & neutral \\ 
\tblexamplesrownumber & Aorta is mildly tortuous and calcified . & the aorta is normal & contradiction \\ 
\bottomrule

\end{tabularx}
\caption{Examples from the development set of MedNLI}
\label{tbl:clinical_nli_random_samples}
\end{table*}

In this paper, we address these gaps and make the following contributions:
\begin{itemize}[noitemsep]
\item Introduce MedNLI -– a new, publicly available, expert annotated dataset for NLI in the clinical domain.
\item A systematic comparison of several state-of-the-art open domain models for NLI on MedNLI.
\item A study of transfer learning techniques from the open domain to the clinical domain. 
\item Techniques for incorporating domain-specific knowledge from knowledge bases (KB) and domain specific data into neural networks.
\end{itemize} 

\section{The MedNLI dataset}
Let us recall the procedure followed for creating the SNLI dataset: Annotators were presented with captions for a Flickr photo (the \textit{premise}) without the photos themselves. They were asked to write three sentences (\textit{hypotheses}): 1) A clearly true description of the photo, 2) A clearly false description, and 3) A description that might be true or false. This procedure produces three training pairs of sentences for each initial premise with three different labels: entailment, contradiction, and neutral, respectively. In order to produce a comparable dataset, we used the same approach, adjusted for the clinical domain. 

\subsection{Premise sampling and hypothesis generation}

As the source of \textit{premise} sentences, we used the MIMIC-III~v1.3~\cite{johnson2016mimic} database. With de-identified records of 38,597 patients, it is the largest repository of publicly available clinical data. Along with medications, lab values, vital signs, etc. MIMIC-III contains 2,078,705 clinical notes written by healthcare professionals in English. The \textit{hypothesis} sentences were generated by clinicians. 

Clinical notes are typically organized into sections such as \texttt{Chief Complaint}, \texttt{Past Medical History, Physical Exam, Impression}, etc. These sections can be easily identified since the associated section headers are often distinctly formatted with capital letters, followed by a colon. The clinicians in our team suggested \texttt{Past Medical History} to be the most informative section of a clinical note, from which critical inferences can be drawn about the patient. 

\begin{figure}
\begin{framed}
\small
You will be shown a sentence from the \texttt{Past Medical History} section of a de-identified clinical note. Using only this sentence, your knowledge about the field of medicine, and common sense:
\begin{itemize}
\item Write one alternate sentence that is \textbf{definitely} a \textbf{true} description of the patient. Example, for the sentence ``Patient has type II diabetes" you could write ``Patient suffers from a chronic condition``
\item Write one alternate sentence that \textbf{might be} a \textbf{true} description of the patient. Example, for the sentence ``Patient has type II diabetes" you could write ``Patient has hypertension"
\item Write one sentence that is \textbf{definitely} a \textbf{false} description of the patient. Example, for the sentence ``Patient has type II diabetes" you could write ``The patient's insulin levels are normal without any medications."
\end{itemize}
\end{framed}
\caption{Prompt shown to clinicians for annotations}
\label{fig:anno-prompt}
\end{figure}

Therefore, we segmented these notes into sections using a simple rule based program capturing the formatting of these section headers. We extracted the \texttt{Past Medical History} section and used a sentence splitter trained on biomedical articles \cite{lingpipe} to get a pool of candidate \textit{premises}. We then randomly sampled a subset from these candidates and presented them to the clinicians for annotation. ~\autoref{fig:anno-prompt} shows the exact prompt shown to the clinicians for the annotation task. SNLI annotations are grounded since they are associated with captions of the same image. We seek to achieve the same goal by grounding the annotations against the medical history of the same patient.

As discussed earlier examples shown in \autoref{tbl:clinical_nli_random_samples} presents unique challenges that involve reasoning over domain-specific knowledge. For instance, the first three examples require the knowledge about clinical terminology. The fourth sample requires awareness medications and the last example elicits knowledge about radiology images. We make the MedNLI dataset available\footnote{\url{https://jgc128.github.io/mednli/}} through the MIMIC-III derived data repository. Thus, any individual certified to access MIMIC-III can access MedNLI.

\subsection{Annotation collection}
Conclusions in the clinical domain are known to be context dependent and a source of multiple uncertainties \cite{han2011varieties}. We had to ensure such subjective interpretations do not result in annotation conflicts affecting the quality of the dataset. To ensure agreement, we worked with clinicians and generated annotation guidelines for a pilot study. Two board certified radiologists worked on the annotation task, and were presented with the 100 unique premises each. 

Some premises, often marred by de-identification artifacts, did not contain any information from which useful inferences could be drawn, e.g. \texttt{This was at the end of [**Month (only) 1702**] of this year.} Such sentences were deemed as invalid for the task and discarded based on clinician judgment. The MIMIC-III dataset contains many de-identification artifacts associated with dates and names (persons and places) which also makes MedNLI more challenging.  

After discarding 16 premises, the result of hypothesis generation was a set of 552 pairs. To calculate agreement, we presented pairs generated by one clinician, and sought annotations from the other clinician, determining if the inference was ``Definitely true", ``Maybe true", or ``Definitely false" \cite{snli}. Comparison of these annotations resulted in a Cohen's kappa of $\kappa=0.78$. While this is substantial if not perfect agreement by itself \cite{mchugh2012interrater}, it is particularly good given the challenging nature of NLI and the complexity of the domain.\footnote{\newcite{rajpurkar2017chexnet} report F1 $<$ 0.45 for four radiologists when compared among themselves} 

On reviewing the annotations, we found that labeling differences between ``Definitely true" and ``Maybe true" were the major source of disagreement. This was primarily because one clinician would think of a scenario that is generally true, while the other would think of assumptions (e.g. patient might be lying, or patient might be pregnant) when it would not.

A discussion with clinicians concluded that the annotation guideline was clear and any person with a formal background of medicine should be able to complete the task successfully. To generate the final dataset, we recruited two additional clinicians, both board certified medical students pursuing their residency programs. Unlike SNLI, we did not collect multiple annotations per sentence pair because of the time and funding constraints. 

\subsection{Dataset statistics}
Together, the four clinicians worked on a total of 4,683 premises over a period of six weeks. The resulting dataset consists of 14,049 unique sentence pairs. 
\begin{table}[ht]
\centering
\begin{tabular}{@{}lr@{}}
\toprule
\textbf{Dataset size}                      & \textbf{} \\
Training pairs                             & 11232     \\
Development pairs                           & 1395      \\ 
Test pairs                                 & 1422      \\ \midrule
\textbf{Average sentence length in tokens} &           \\
Premise                                    & 20.0      \\
Hypothesis                                 & 5.8       \\ \midrule
\textbf{Maximum sentence length in tokens} &           \\
Premise                                    & 202      \\
Hypothesis                                 & 20       \\ \bottomrule
\end{tabular}
\caption{Key statistics of the dataset}
\label{tbl:clinical_nli_statistics}
\end{table}
\begin{figure}[ht]
\begin{center}
\centering
    \includegraphics[width=0.95\linewidth, clip, trim=0.2cm 0.2cm 0.2cm 0.2cm]{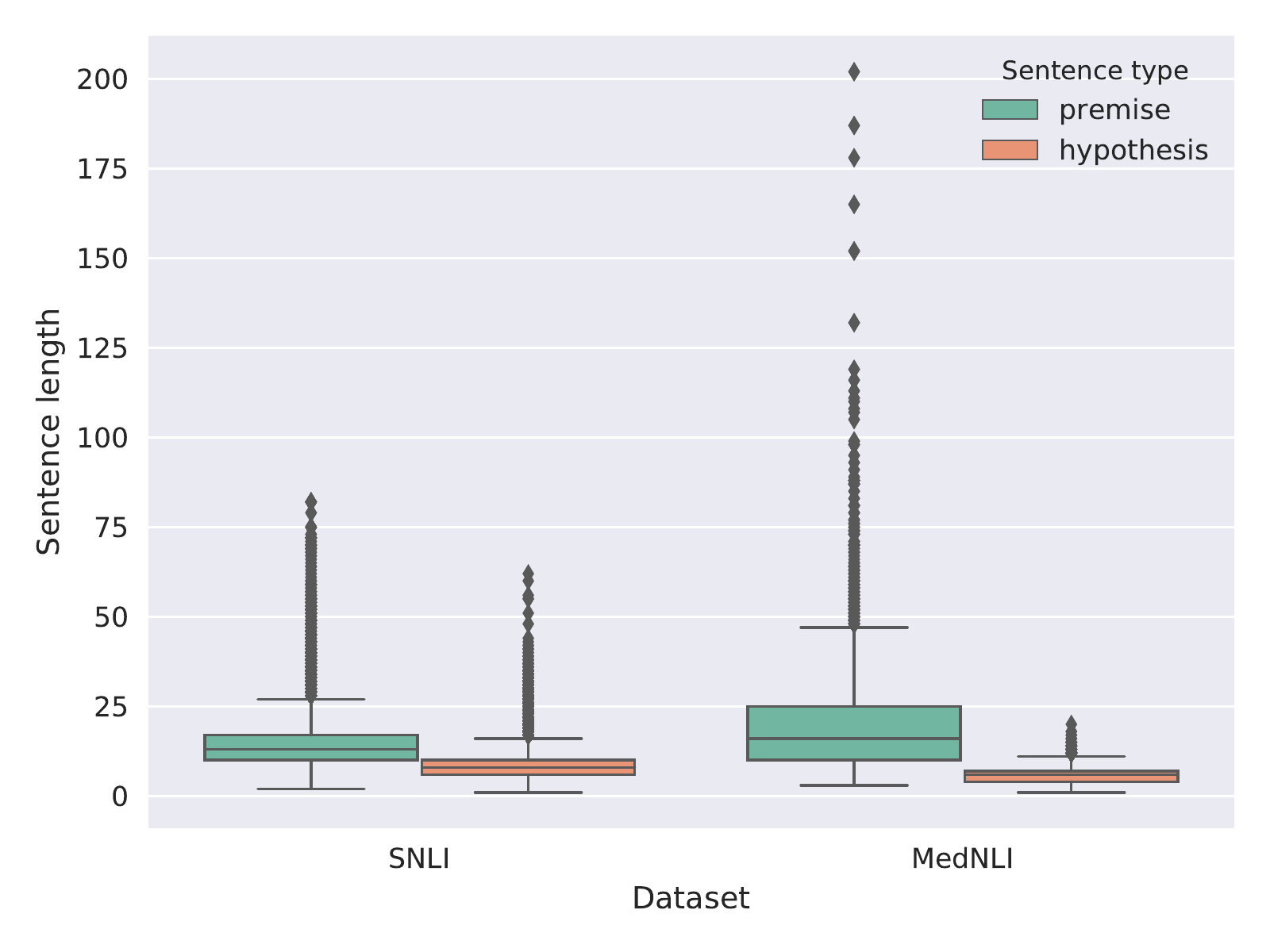}
\caption{Box plot of the distribution of sentence length in tokens in SNLI and MedNLI}
\label{fig:snli_clinical_nli_sentence_len_distributions}
\end{center}
\end{figure}
Following \newcite{snli}, we split the dataset into training, development, and testing subsets and ensured that no premise was overlapping between the three subsets. \autoref{tbl:clinical_nli_statistics} presents key statistics of MedNLI, and
the~\autoref{fig:snli_clinical_nli_sentence_len_distributions} shows the distribution of the lengths of the sentences compared to the SNLI dataset. Note that, similar to SNLI, the premises, on average, are longer than the hypotheses and have more variation in length. 
\begin{table*}[ht]
\centering
\begin{tabular}{l l r}
\toprule
\textbf{Semantic type} & \textbf{Common examples}       & \multicolumn{1}{l}{\textbf{Count}} \\ \midrule
 Finding & \textit{asymptomatic, history of kidney stones, nystagmus } & 35,439 \\
Disease or Syndrome & \textit{chf, enterovaginal fistula, diverticulitis, acute stroke} & 9,941 \\
Sign or Symptom  & \textit{chest pain, dyspnea, seizures, vomiting, nausea} & 5,294 \\
Therapeutic Procedure & \textit{aspiration, cabg, limb perfusion, chemotherapy} & 5,043 \\
Pharmacological Substance & \textit{lopressor, morphine, atenolol, ativan, coumadin} & 3,948 \\
Body part, organ & \textit{r arm, jaw, left frontal lobe brain, patellar tendon} & 3,907 \\
Laboratory Procedure & \textit{serum glucose, blood ph, cbc, hematocrit, neutrophil count } & 1,136 \\
\bottomrule
\end{tabular}
\caption{Examples of medical concepts belonging to common semantic types across premises and hypotheses in the MedNLI training data.}
\label{tbl:semtypes_stats}
\end{table*}

Medical concepts expressed in MedNLI belong to several categories such as medications, diseases, symptoms, devices, etc. We used Metamap \cite{aronson2010overview} -- a tool to identify medical concepts from text and map them to standard terminologies in the Unified Medical Language System (UMLS) \cite{bodenreider2004unified}. Further, the UMLS can be queried to identify the \textit{Semantic Type} of each medical concept. There are 133 semantic types in the UMLS ranging from \textit{Finding, Medical Device} to \textit{Bacteria} and \textit{Hormone}.\footnote{\url{https://metamap.nlm.nih.gov/SemanticTypesAndGroups.shtml}} \autoref{tbl:semtypes_stats} shows common occurrences of medical concepts belonging to a few clinical semantic types in the train set. These examples show that medical concepts can be abbreviations or multiword expressions and comprise of a vocabulary that is very different from the open domain.

\section{Models}
\label{sec:models}
To establish a baseline performance on MedNLI, we experimented with a feature-based system. To further explore the performance of modern neural networks-based systems, we experimented several models of various degrees of complexity: Bag of Words (BOW), InferSent~\cite{conneau2017supervised}
and ESIM~\cite{chen2017enhanced}. Note that our goal here is not to outperform existing models, but to explore the relative gain of the proposed methods, and compare them to a baseline. We used the same set of hyperparameters in all models to ensure that any difference in performance is exclusively due to the algorithms.

\paragraph{Feature-based system} We used a gradient boosting classifier incorporating a variety of hand crafted features. Apart from standard NLP features, we also infused clinical knowledge from the Unified Medical Language System (UMLS)~\cite{bodenreider2004unified}. Each terminology in the UMLS can be viewed as a graph where nodes represent medical concepts, and edges represent relations between them. These are canonical relationships found in ontologies such as \textit{IS A} and \textit{SYNONYMY}. For instance, \textit{diabetes} \textit{IS A} \textit{disorder of the endocrine system}. The domain specific features we added to the model represent similarity between UMLS concepts from the premise and the hypothesis, based how close they appear in the UMLS graph \cite{pedersen2007measures}. Following \cite{shivade2015textual,pedersen2007measures} we used the SNOMED-CT terminology in our experiments.

The groups below summarize the feature sets used in our model (35 features in total):
\begin{enumerate}[noitemsep]
    \item BLEU score
    \item Number of tokens (e.g. min, max, difference)
    \item Negations (e.g. keywords such as \textit{no, do not})
    \item TF-IDF similarity (e.g. cosine, euclidean)
    \item Edit distances (e.g. Levenshtein)
    \item Embedding similarity (e.g. cosine, euclidean)
    \item UMLS similarity features (e.g. shortest path distance between UMLS concepts)
\end{enumerate}

\paragraph{BOW model.} We use a bag-of-words (BOW) model as a simple baseline for the NLI task: the \textit{Sum of words} model by \newcite{snli} with a small modification. While \newcite{snli} use $tanh$ as the activation function in the model, we use $\relu$, since it trained faster and achieved better results~\cite{glorot2011deep}. 
In order to represent an input sentence as a single vector, this architecture simply sums up the vectors of individual tokens. The premise and hypothesis vectors are then concatenated and passed through a multi-layer neural network. Recent work shows that even this straightforward approach encodes a non-trivial amount of information about the sentence~\cite{adi2016fine}.

\paragraph{InferSent model.} InferSent \cite{conneau2017supervised} is a model for sentence representation that demonstrated close to state-of-the-art performance across a number of tasks in NLP (including NLI) and computer vision. The main differences from the BOW model are as follows: 
\begin{itemize}[noitemsep]
\item A bidirectional LSTM encoder of input sentences and a max-pooling operation over timesteps are used to get a vector for the premise ($p$) and for the hypothesis ($h$); 
\item A more complex scheme of interaction between the vectors $p$ and $h$ to get a single vector $z$ that contains all the information needed to produce a decision about the relationship between the input sentences: $z = [p,h, |p-h|, p*h]$. 
\end{itemize}

\begin{figure}[ht]
  \centering
  \includegraphics[width=0.6\linewidth]{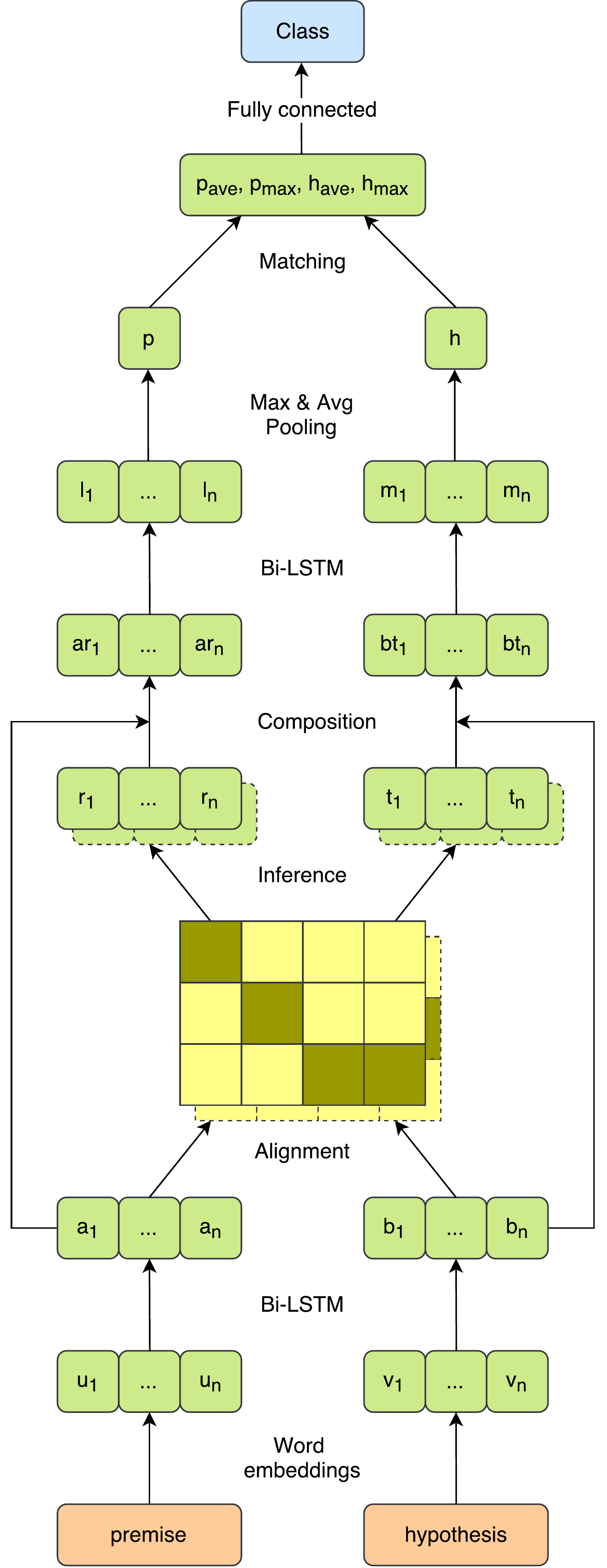}
  \caption{ESIM model. Dashed blocks illustrate the knowledge-directed attention matrix and the corresponding vectors (see Section~\ref{sec:knowledge_directed_attention} for details).}
  \label{fig:models_esim}
\end{figure}

\paragraph{ESIM model.} The ESIM model, developed by~\newcite{chen2017enhanced}, is shown in~\autoref{fig:models_esim}. It is a fairly complex model that makes use of two bidirectional LSTM networks. The basic idea of ESIM is as follows: 

\begin{itemize}[noitemsep]
    \item The first LSTM produces a sequence of hidden states.
    \item Pairwise attention matrix $e$ is computed between all tokens in the premise and the hypothesis to produce new sequences of ``attended'' hidden states, which are then fed into the second LSTM. 
    \item Max and average pooling are performed over the output of the LSTMs. 
    \item The output of the pooling operations is combined in a way similar to the InferSent model.
\end{itemize}

The three aforementioned models exemplify the architectures that are, perhaps, the most widely used for NLI task, spanning from simple bag-of-words approaches to complicated models with Bi-LSTM and inter-sentence attention. We additionally experimented with a plain Bi-LSTM model as well as GRU~\cite{cho2014properties}, but since their performance was not remarkable (in the same range as BOW) we do not report it here.
\subsection{Transfer learning}

Given the existence of larger general-domain NLI datasets such as SNLI and MultiNLI, it stands to reason to try to leverage them to improve the performance in the clinical domain. Transfer learning has been shown to improve performance on variety of tasks such as: machine translation on low-resource languages~\cite{zoph2016transfer} and also some tasks from the bio-medical domain in particular~\cite{sahu2017matters,lee2017transfer}. To see if a corresponding boost would be possible for the NLI task, we investigated three common transfer learning techniques on the MedNLI dataset using SNLI and five different genres from MultiNLI.

\textbf{Direct transfer} is the simplest method of transfer learning. After training a model on a large \textit{source domain} dataset, the model is directly tested on the \textit{target domain} dataset. If the source and the target domains are similar to some extent, one can achieve a reasonable accuracy by simply applying a model pre-trained on the source domain to the target domain. In our case the source domain is general domain in SNLI and the various genres in MultiNLI, and the target domain is clinical. 

\textbf{Sequential transfer} is the most widely used technique. After pre-training the model on a large source domain, the model is further fine-tuned using the smaller training data of the target domain. The assumption is that while the model would learn domain-specific features, it would also learn some domain-independent features that will be useful for the target domain. Furthermore, the fine-tuning process would affect the learned features from the source domain and make them more suitable for the target domain. 

\begin{figure}
  \centering
  \includegraphics[width=0.5\linewidth]{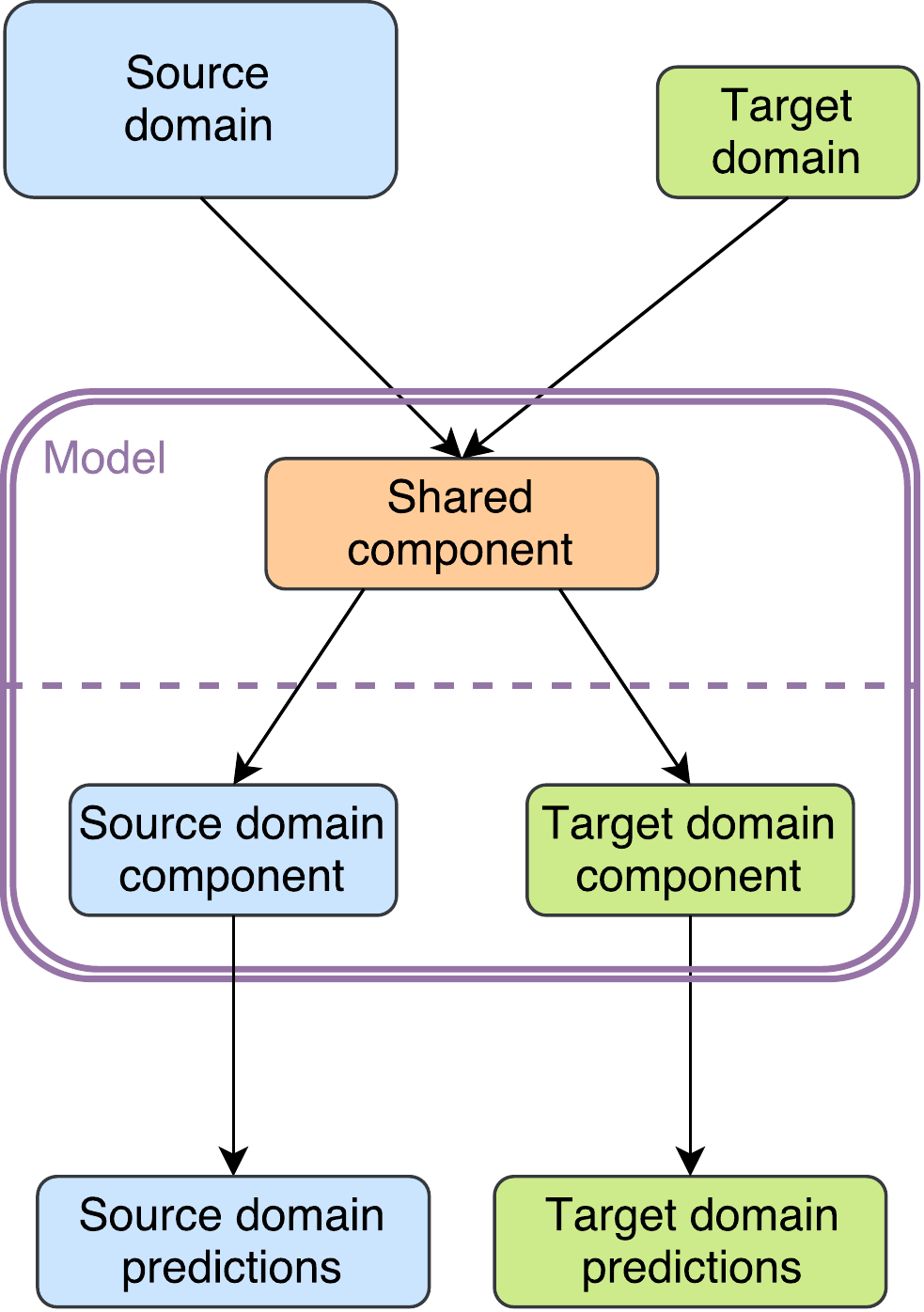}
  \caption{Schematic depiction of the model for multi-target transfer learning}
  \label{fig:transfer_multi_trarget}
\end{figure}

\textbf{Multi-target transfer} is a more complex method (see~\autoref{fig:transfer_multi_trarget}). It involves separation of the model into three components (or layers):
\begin{itemize}[noitemsep]
    \item \textit{The shared component} is trained on both the source and target domains;
    \item \textit{The source domain component} is trained only during the pre-training phase and does not participate in the prediction of the target domain;
    \item \textit{The target domain component} is trained during the fine-tuning stage and it produces the predictions together with the shared component.
\end{itemize}

The motivation for multi-target transfer is that the performance should improve by splitting deeper layers of the model into domain-specific parts and having a shared block early in the network, where it presumably learns domain-independent features. The target-specific component will not be in the local minimum of the source domain after the pre-training stage, enabling the model to find a better local minimum for the target domain.

\subsection{Word embeddings}
Another way to improve the accuracy on the target domain is to use domain-specific word embeddings instead of, or, in addition to, open-domain ones. For example, \newcite{stanovsky2017recognizing} achieved state of the art results in recognizing Adverse Drug Reaction using graph-based embeddings trained on the ``Drugs'' and ``Diseases'' categories from DBpedia~\cite{lehmann2015dbpedia}, as well as embeddings trained on web-pages categorized as ``medical domain''.

We experimented  with the following publicly available general-domain word embeddings:
\begin{itemize}[noitemsep]
    \item \textbf{GloVe\textsubscript{[CC]}}: GloVe embeddings~\cite{pennington2014glove}, trained on Common Crawl\footnote{\url{http://commoncrawl.org/}}.
    \item \textbf{fastText\textsubscript{[Wiki]}}: fastText embeddings~\cite{bojanowski2016enriching}, trained on Wikipedia.
    \item \textbf{fastText\textsubscript{[CC]}}: fastText embeddings, trained on Common Crawl.
\end{itemize}

Furthermore, we trained fastText embeddings on the following domain-specific corpora:
\begin{itemize}[noitemsep]
    \item \textbf{fastText\textsubscript{[BioASQ]}}:A collection of PubMed abstracts from the BioASQ challenge data~\cite{tsatsaronis2015overview}. This data includes abstracts from 12,834,585 scientific articles from the biomedical domain.
    \item \textbf{fastText\textsubscript{[MIMIC-III]}}: Clinical notes for patients from the MIMIC-III database~\cite{johnson2016mimic}: 2,078,705 notes with 320 tokens in each on average.
\end{itemize}

Finally, we experimented with initializing word embeddings with pre-trained vectors from general domain and further training on a domain-specific corpus:
\begin{itemize}[noitemsep]
    \item GloVe\textsubscript{[CC]} \textrightarrow{} fastText\textsubscript{[BioASQ]}: GloVe embeddings for initialization, and the BioASQ data for fine-tuning.
    \item GloVe\textsubscript{[CC]} \textrightarrow{} fastText\textsubscript{[BioASQ]} \textrightarrow{} fastText\textsubscript{[MIMIC-III]}: GloVe embeddings for initialization, and two consequent fine-tuning using the BioASQ and MIMIC-III data.
    \item fastText\textsubscript{[Wiki]} \textrightarrow{} fastText\textsubscript{[MIMIC-III]}: fastText Wikipedia embeddings for initialization, and the MIMIC-III data for fine-tuning.
\end{itemize}

Experiments using other approaches to word embeddings, such as \texttt{word2vec}~\cite{mikolov2013distributed} and  CoVe~\cite{McCann2017LearnedIT} did not show any gains. All the word embeddings used in this work are available for download. \footnote{\url{https://jgc128.github.io/mednli/}}


\subsection{Knowledge integration}
Since processing of medical texts requires domain-specific knowledge, we experimented with different ways of incorporating such knowledge into the systems. First, we can modify the input to the system so it carries a portion of clinical information. Second, we can modify the model itself, integrating domain knowledge directly into it. 

The UMLS, is the largest, publicly available, and regularly updated database of medical terminologies, concepts, and relationships between them. It can be viewed as a graph where clinical concepts are nodes, connected by edges representing relations, such as synonymy, parent-child, etc. 
Following past work \cite{pedersen2007measures}, we restricted to the SNOMED-CT terminology in UMLS and experimented with two techniques for incorporating knowledge: retrofitting and attention.

\subsubsection{Retrofitting}
\textit{Retrofitting}~\cite{faruqui:2014:NIPS-DLRLW} modifies pre-trained word embeddings based on an ontology. The basic idea is to try to bring the representations of the concepts that are connected in the ontology closer to one another in vector space. The authors showed that retrofitting using WordNet~\cite{wordnet} synsets improves accuracy on several word-level tasks, as well as on the sentiment analysis task.

\subsubsection{Knowledge-directed attention}
\label{sec:knowledge_directed_attention}
Attention proved to be a useful technique for many NLP tasks, starting from machine translation~\cite{bahdanau2014neural} to parsing~\cite{vinyals2015grammar} and NLI itself~\cite{parikh2016decomposable,rocktaschel2015reasoning}. In most models (including the ESIM model that we use in our experiments) attention is learned in an end-to-end fashion. However, if we have knowledge about relationships between concepts, we could leverage it to explicitly tell the model to attend to specific concepts during the processing of the input sentence.

For example, there is an edge in SNOMED-CT from the concept \textit{Lung consolidation} to \textit{Pneumonia}. Using this information, during the processing of a sentence pair 
\begin{itemize}[noitemsep]
    \item \textbf{Premise} The patient has \textit{pneumonia}.
    \item \textbf{Hypothesis} The patient has a \textit{lung} disease.
\end{itemize}
the model could attend to the token \textit{lung} while processing \textit{pneumonia}. 

We propose to integrate this knowledge in a way similar to how attention is used in the ESIM model. Specifically, we calculate the attention matrix $e \in \mathbb{R}^{n \times m} $ between all pairs of tokens $a_i$ and $b_j$ in the inputs sentences, where $n$ is the length of the hypothesis and $m$ is the length of the premise. The value in each cell reflects the length of the shortest path $l_{ij}$ between the corresponding concepts of the premise and the hypothesis in SNOMED-CT. 



This process could be informally described as follows: each token $\tilde{a_i}$ of the premise is a weighted sum of relevant tokens $b_j$ of the hypothesis, according to the medical ontology, and vice versa. This enables the medical domain knowledge to be integrated directly into the system. 

We used the original tokens $a_i$ as well as the attended $\tilde{a_i}$ inside the model for both InferSent and ESIM. For InferSent, we simply concatenate them across the time dimension: $$\hat{a} = [a_1, a_2, \dots, a_n, \tilde{a_1}, \tilde{a_2}, \dots, \tilde{a_n}]$$ where $n$ is the length of the inputs sequence. For the ESIM model, we concatenate $a_i$ and $\tilde{a_i}$ before passing them to the composition layer (see~\autoref{fig:models_esim} and Section~3.3 in the original paper~\cite{chen2017enhanced}). This enables the model to learn the relative importance of both the token and the knowledge directed attention.

\begin{table}[ht]
\centering
\begin{tabular}{@{}lrrrr@{}}
\toprule
\multicolumn{1}{l}{\textbf{Set}} & \multicolumn{1}{c}{\textbf{Features}} & \multicolumn{1}{c}{\textbf{BOW}} & \multicolumn{1}{c}{\textbf{InferSent}} & \multicolumn{1}{c}{\textbf{ESIM}} \\ \midrule
Dev     & 51.9  & 71.9      & \textbf{76.0}      & 74.4                            \\
Test    & 51.9  & 70.2      & \textbf{73.5}      & 73.1                            \\ \bottomrule
\end{tabular}%
\caption{Baseline accuracy on the development and the test set of MedNLI for different models.}
\label{tbl:results_baseline}
\end{table}

\section{Results and discussion}

\begin{table*}[ht]
\centering

\resizebox{0.9\textwidth}{!}{%

\begin{tabular}{@{}lrrrrrrrrr@{}}
\toprule
\multirow{2}{*}{\textbf{Source domain}} & \multicolumn{3}{l}{\textbf{Direct transfer}} & \multicolumn{3}{l}{\textbf{Sequential transfer}} & \multicolumn{3}{l}{\textbf{Multi-target transfer}} \\ \cmidrule(l){2-10} 
 & \multicolumn{1}{l}{\textbf{BOW}} & \multicolumn{1}{l}{\textbf{InferSent}} & \multicolumn{1}{l}{\textbf{ESIM}} & \multicolumn{1}{l}{\textbf{BOW}} & \multicolumn{1}{l}{\textbf{InferSent}} & \multicolumn{1}{l}{\textbf{ESIM}} & \multicolumn{1}{l}{\textbf{BOW}} & \multicolumn{1}{l}{\textbf{InferSent}} & \multicolumn{1}{l}{\textbf{ESIM}} \\ \midrule
snli & -21.8 & \textbf{-24.2} & -22.8 & 1.8 & -1.8 & -2.5 & \textbf{2.4} & -2.5 & -0.7 \\
fiction & \textbf{-21.6} & -25.6 & \textbf{-21.4} & 1.3 & 0.4 & -0.5 & 1.4 & 0.1 & \textbf{0.3} \\
government & -23.8 & -27.2 & -26.2 & 1.0 & 0.8 & -0.7 & 1.3 & 0.2 & 0.2 \\
slate & -23.2 & -25.7 & -21.6 & \textbf{1.9} & \textbf{0.9} & \textbf{-0.2} & 1.1 & \textbf{0.6} & -0.1 \\
telephone & -25.7 & -27.3 & -25.6 & 1.7 & -0.2 & -1.1 & 1.2 & 0.4 & -0.1 \\
travel & -25.4 & -29.1 & -23.5 & 1.6 & 0.0 & -0.7 & 0.2 & -0.3 & 0.1 \\ \bottomrule
\end{tabular}

}
\caption{Absolute gain in accuracy with respect to the baseline (see~\autoref{tbl:results_baseline}) on the MedNLI test set for different transfer learning modes. Bold indicates the best source domain for each model and transfer.}
\label{tbl:results_transfer_learning}
\end{table*}

\subsection{Implementation details}
For the features-based system we used the \verb|GradientBoostingClassifier| from the scikit-learn library \cite{scikit-learn}. We implemented all models using PyTorch\footnote{\url{https://pytorch.org/}} and trained them with the Adam optimizer~\cite{kingma2014adam} until the validation loss showed no improvement for 5 epochs. The epoch with the lowest loss on the validation set was selected for testing. We used the GloVe word embeddings~\cite{pennington2014glove} in all experiments, except for~\autoref{sec:results_word_embeddings}. In all experiments we report the average result of 6 different runs, with the same hyperparameters and different random seeds. Medical concepts in SNOMED-CT were identified in the premise and hypothesis sentences using Metamap \cite{aronson2010overview}. The code for all experiments is publicly available.\footnote{\url{https://jgc128.github.io/mednli/}}

\subsection{Baselines}
\autoref{tbl:results_baseline} shows the baseline results: the performance of a model when trained and tested on the MedNLI dataset. The feature-based system performed the worst. As for neural networks-based systems, the BOW model showed the lowest performance on the both development and test sets. The InferSent model, in contrast, achieved the highest accuracy, despite ESIM outperforming it on SNLI. This could be attributed to the fact that ESIM has twice as many parameters as InferSent, and so InferSent overfits less to the smaller MedNLI dataset.

\subsection{Transfer learning}

As expected, ~\autoref{tbl:results_transfer_learning} shows that direct transfer is worse than the baseline but is still better than a random baseline of 33.3\%. Sequential and multi-target transfer learning, in contrast, yields a considerable gain for all the models. The maximum gain is 2.4\%, 0.9\%, and 0.3\% for the BOW, InferSent, and ESIM models correspondingly. 

Second, note that the biggest SNLI domain gave the most boost in only two out of six cases, implying that the size of the domain should not be the most important factor in choosing the source domain for transfer learning. The best accuracy for all the models was obtained with the ``slate'' domain from MultiNLI corpus with sequential transfer (note, however, that the accuracy of ESIM is actually lower than the baseline accuracy). This is consistent with observations of \newcite{multinli}. Finally, although some domains are better for particular transfer learning methods with particular models, there is no single combination that works for all cases. 

\subsection{Word embeddings}
\label{sec:results_word_embeddings}
~\autoref{tbl:results_word_embeddings} shows that simply using of the embeddings trained on the MIMIC-III notes significantly increases the accuracy for all the models. Furthermore, the InferSent models achieves a 3.1\% boost with the fastText Wikipedia embeddings, fine-tuned on the MIMIC-III data. Note that the results fastText\textsubscript{[Wiki]} are worse than the baseline GloVe\textsubscript{[CC]} for all models, which could be due to the source corpus size. However, the results on BioASQ are worse than on MIMIC-III, despite the significantly larger corpus of the BioASQ embeddings. 
Overall, our experiments show the benefit of domain-specific rather than general-domain word embeddings.

\begin{table}[ht]
\centering
\resizebox{1\linewidth}{!}{%

\begin{tabular}{@{}lrrr@{}}
\toprule
\multicolumn{1}{c}{\textbf{Embeddings}}              & \multicolumn{1}{c}{\textbf{BOW}} & \multicolumn{1}{c}{\textbf{InferSent}} & \multicolumn{1}{c}{\textbf{ESIM}} \\ \midrule
fastText\textsubscript{[Wiki]}                                             & -3.5         & -3.5               & -4.4          \\
fastText\textsubscript{[CC]}  (600B)                                   & -0.6         & 1.3                & -0.3          \\
fastText\textsubscript{[BioASQ]}  (2.3B)                                         & 0.5          & 0.6                & 0.2           \\
fastText\textsubscript{[MIMIC-III]}  (0.8B)                                      & \textbf{1.1} & 2.3                & 1.2           \\
GloVe\textsubscript{[CC]} \textrightarrow{} fastText\textsubscript{[BioASQ]}                       & 0.2          & 0.7                & 1.4           \\
GloVe\textsubscript{[CC]} \textrightarrow{} fastText\textsubscript{[BioASQ]} \textrightarrow{} fastText\textsubscript{[MIMIC-III]} & 0.9          & 2.7                & \textbf{1.8}  \\
fastText\textsubscript{[Wiki]} \textrightarrow{} fastText\textsubscript{[MIMIC-III]}                      & 0.1          & \textbf{3.1}       & 1.7           \\ \bottomrule
\end{tabular}%

}
\caption{Absolute gain in accuracy with respect to the baseline (GloVe\textsubscript{[CC]}) for different word embeddings (the number in parentheses reflects the number of tokens in the corresponding training corpora).}
\label{tbl:results_word_embeddings}
\end{table}

To improve the accuracy even further, we explore a common way of improving predictions of machine learning models: merging of the predictions of several independent models in an ensemble. Ensemble models achieve better performance when base classifiers have complementary strengths. We combined by summation the predictions of the six models described in the previous paragraph, and the resulting ensemble achieved a small gain of 0.4\% in accuracy compared to the best base models. This is a simple way of improving the accuracy that does not require any additional data or training, except for the models themselves.

\subsection{Knowledge integration}
\subsubsection{Retrofitting}
~\autoref{tbl:results_retrofitting} shows that retrofitting only hurts the performance. This is in contrast with the results of the original study, where retrofitting was beneficial not only for word-level tasks but also for tasks such as sentiment analysis~\cite{faruqui:2014:NIPS-DLRLW}. We hypothesize that although WordNet and UMLS are structurally similar, significant differences in the content ~\cite{burgun2001comparing} might be the reason for these results. Retrofitting should be more useful when it is used on a WordNet-like database where the main relation is synonymy, and tested on tasks such as word similarity tests or sentiment analysis. 
The UMLS semantic network is more complex and contains relations that may not be suitable for retrofitting.

\begin{figure}[ht]
\begin{center}
\centering

\begin{tikzpicture}
    \node[anchor=south west,inner sep=0] (image) at (0,0) {
        \includegraphics[width=0.80\linewidth, clip, trim=0.2cm 0.2cm 0.2cm 0.2cm]{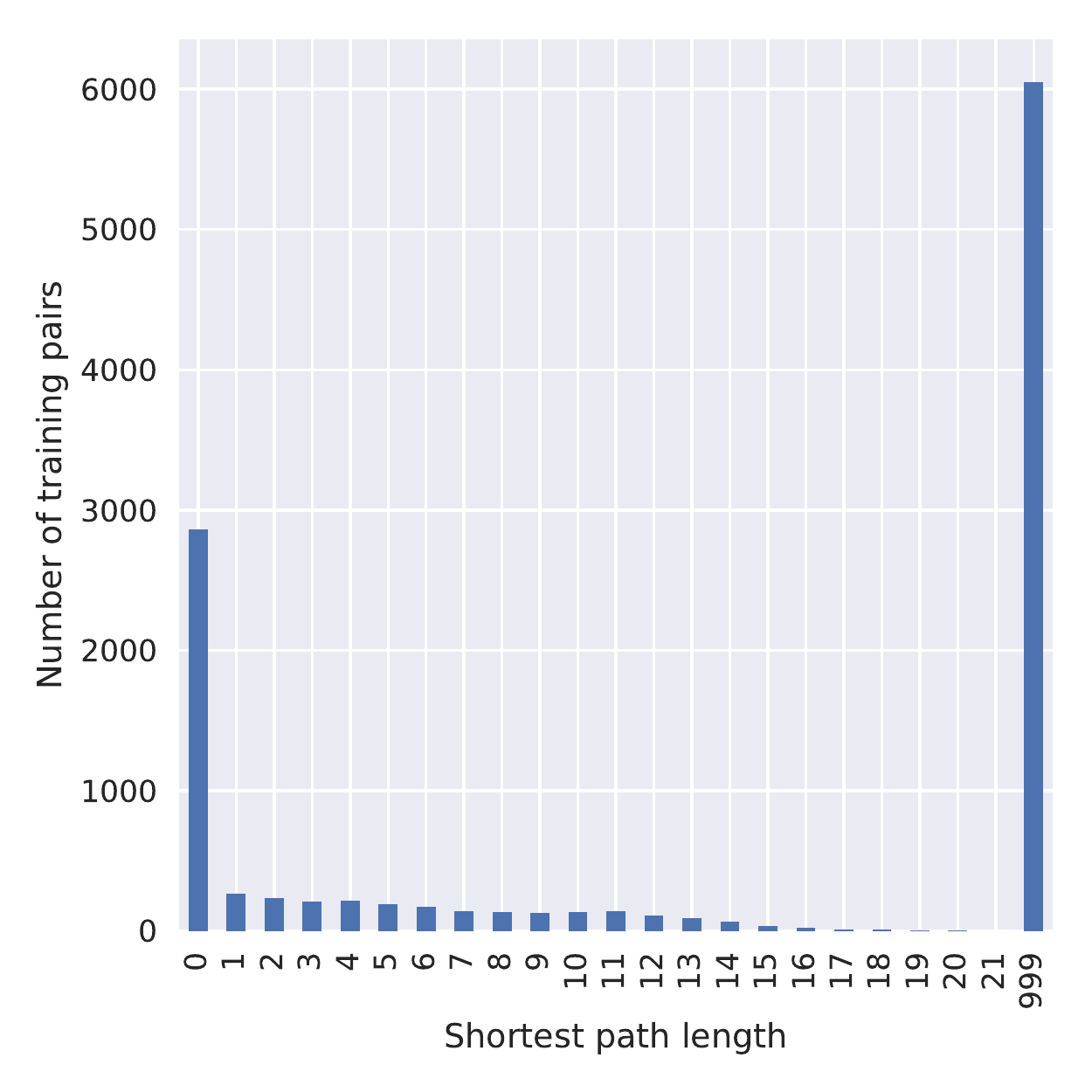}
    };
    \begin{scope}[x={(image.south east)},y={(image.north west)}]
        \fill [white] (0.935,0.05) rectangle (0.98,0.12);
        \node [rotate=90] at (0.965,0.075) {{\fontsize{4}{6}\selectfont No path}};
    \end{scope}
\end{tikzpicture}

\caption{Lengths of the shortest paths between concepts in the premise and the hypothesis. $0$ indicates that they contain the same concept.
}
\label{fig:shortest_path_length_clinical}
\end{center}
\end{figure}

Moreover, retrofitting works only on directly related concepts in a knowledge graph (although it might affect, to some extent, indirectly related concepts by transitivity). However, \autoref{fig:shortest_path_length_clinical} shows that very few training pairs have such concepts (namely, pairs with a path of length 1). In contrast, the lengths of the shortest path in SNLI using WordNet fall close to 1. 
This suggests that the medical inferences represented in MedNLI requires more complex reasoning, typically involving multiple steps.

As a sanity check, we applied retrofitting to the GloVe embeddings and tested the InferSent model on the ``fiction'' domain from the MultiNLI corpus. We used the code and lexicons provided by \newcite{faruqui:2014:NIPS-DLRLW} and confirmed that retrofitting hurts the performance in that case as well. 

\begin{table}[ht]
\centering
\begin{tabular}{@{}lll@{}}
\toprule
\textbf{BOW} & \textbf{InferSent} & \textbf{ESIM} \\ \midrule
-1.7 & -2.0 & -2.7 \\ \bottomrule
\end{tabular}
\caption{Absolute gain in accuracy using retrofitting for MedNLI.}
\label{tbl:results_retrofitting}
\end{table}

\subsubsection{Knowledge-directed attention}

To evaluate the potential of knowledge-directed attention, let us consider its effect on a baseline embedding (GloVe\textsubscript{[CC]}) and a fastText embedding trained on MIMIC-III (fastText\textsubscript{[MIMIC-III]}) that showed good performance in section \ref{sec:results_word_embeddings}.

Knowledge-directed attention showed positive effect with the InferSent model on GloVe\textsubscript{[CC]} (0.3 gain), and was not detrimental to ESIM. However, in case of the fastText\textsubscript{[MIMIC-III]} embeddings knowledge-directed attention was beneficial to both models, as shown in~\autoref{tbl:results_attention}. 
Note that while retrofitting can use only direct relations during the training process, our method incorporates information about relationships of any length, which is a necessity (as evident from~\autoref{fig:shortest_path_length_clinical}). 


\begin{table}[ht]
\centering
\begin{tabular}{@{}lrr@{}}
\toprule
\textbf{Embedding} & \multicolumn{1}{l}{\textbf{InferSent}} & \multicolumn{1}{l}{\textbf{ESIM}} \\ \midrule

GloVe\textsubscript{[CC]}          & 0.3     & 0.0   \\
fastText\textsubscript{[MIMIC-III]}      & 0.2     & 0.3   \\ \bottomrule

\end{tabular}
\caption{Absolute gain in accuracy using knowledge-directed attention.}
\label{tbl:results_attention}
\end{table}

\begin{table*}[ht]
\centering

\begin{tabularx}{\linewidth}{l X X l l}
\toprule
\textbf{Category} & \textbf{Premise}                                                                                                                           & \textbf{Hypothesis}                                                   & \textbf{Predicted} & \textbf{Expected} \\ \midrule

Numerical & On weaning to 6LNC, his O2 decreased to 81-82\%  & He has poor O2 stats & neutral & entailment \\
Reasoning & WBC 12 , Hct 41 .
 & WBC slightly elevated & contradiction & entailment \\
 \midrule
 World & The infant emerged with spontaneous cry. & The infant was still born. & entailment & contradiction \\
 Knowledge & No known sick contacts & No recent travel & entailment & neutral \\
 \midrule
 Abbreviation & No CP or fevers. & Patient has no angina & neutral & entailment \\
 & Received GI cocktail for h/o GERD,  esophageal spasm & Received a proton pump inhibitor & entailment & neutral \\
 \midrule
 Medical & EKG showed T-wave depression in V3-5, with no prior EKG for comparison. & Patient has a normal EKG & neutral & contradiction \\
Knowledge & Mother developed separation of symphysis pubis and was put in traction .
& She has orthopedic injuries & neutral & entailment \\
\midrule
Negation & Head CT was negative for bleed. & The patient has intracranial hemorrhage & neutral & contradiction \\
 & Denied headache, sinus tenderness, or congestion & Patient has headaches & neutral  & contradiction \\
\bottomrule
\end{tabularx}
\caption{Representative errors made by different models}
\label{tbl:errors}
\end{table*}

\section{Discussion}
 \subsection{Error analysis} 
 The neutral class is the hardest to recognize for all models and their modifications. Majority errors stem from confusion between entailment and the neutral class. Use of domain-specific embeddings trained on MIMIC-III result in gains which are equally distributed across all three classes. Interestingly, gains from knowledge-directed attention stem mostly (60\%) from the neutral class. Moreover, 87\% of these neutral predictions were predicted as entailment before adding the knowledge directed attention.
 
 We categorized the errors made by all the models in four broad categories. \autoref{tbl:errors} outlines representative errors made by most models in these categories. Numerical reasoning such as \textit{abnormal lab value} $\rightarrow$ \textit{disease} or \textit{abnormal vital sign} $\rightarrow$ \textit{finding} are very hard for a model to learn unless it has seen multiple instances of the same numerical value.\footnote{The symbol $\rightarrow$ represents entailment relationship} The first step is to learn what values are \textit{abnormal} and the next is to actually perform the inference. This has been identified as a major challenge for NLI since long \cite{sammons2010ask}. Many inferences require world knowledge that could be deemed close to open domain NLI . While these are very subtle, some are quite domain specific (e.g. \textit{emergency admission} $\nrightarrow$ \textit{planned visit}). Abbreviations are ubiquitously found in clinical text. While some are standard and therefore frequent, clinicians tend to use non standard abbreviations making inference harder. Finally, many inferences are at the core of reasoning with clinical knowledge. While training on large datasets maybe a natural but impractical solution, this is an open research problem for researchers in the community.
 
 Following Multi-NLI \cite{nangia2017repeval} we also probed for prediction patterns with linguistic features like active-passive voice, negations, temporal expressions, coreference and modal verbs. As is common with tasks in clinical NLP, negations play a crucial role in NLI. All three models interpret negations correctly to a large extent (upto 75\%). Besides negation, other linguistic features have a sparse presence in MedNLI and none indicated a subset that is significantly harder. We did not identify any qualitative pattern in the gains resulting from adding domain knowledge through word embeddings or knowledge-directed attention.
 
 We also conducted a systematic error analysis with respect to the semantic types (recall \autoref{tbl:semtypes_stats}) of concepts found in the premise and hypothesis pairs. We had hoped to find patterns such as \textit{Finding} \textrightarrow \textit{Pharmacological Substance} or \textit{Sign or Symptom} \textrightarrow \textit{Disease or Syndrome} but nothing stood out. A significant number of instances where models were correct, involved medical concepts that were either a \textit{Finding} or \textit{Disease or Syndrome}. However, this maybe simply because of their larger presence in the data than other semantic types in general. 
 
\subsection{Limitations}
Unlike SNLI and MultiNLI, each example in the MedNLI dataset was single annotated. However, this was the best we could do in the limited time and resources available. Very recently \newcite{gururangan2018annotation} discovered presence of annotation artifacts in NLI datasets. Similar findings have also been reported by \newcite{tsuchiya2018performance} and \newcite{poliak2018hypothesis}. Since we followed the exact same process, we found artifacts to be present in MedNLI as well. The premise-oblivious text-classifier from \newcite{gururangan2018annotation} that achieves an F1 of 67.0 on SNLI, and 53.9 on Multi-NLI achieves 61.9 on MedNLI.

\subsection{Applications}Finally, development of NLI for the clinical domain has exciting real world applications. Condition criteria are one of the key primitives in clinical practice guidelines and clinical trials \cite{weng2010formal}. These criteria expressed as natural language sentences are ``based on such factors as age, gender, the type and stage of a disease, previous treatment history, and other medical conditions".\footnote{\url{http://clinicaltrials.gov}} Fulfillment of eligibility criteria for research studies such as clinical trials can be automated using NLI \cite{shivade2015textual}. Each eligibility criterion sentence can be treated as a premise and each sentence in the clinical note as a hypothesis. If reading a sentence in the clinical note of a patient allows a model to conclude that the criterion is entailed, the patient is said to satisfy the criterion. Similar application can be sought for monitoring clinical guideline compliance where each guideline statement can be treated as a premise.

\subsection{Conclusion}
We have presented MedNLI, an expert annotated, public dataset for natural language inference in the clinical domain. To the best of our knowledge, MedNLI is the first dataset of its kind. Our experiments with several state-of-the-art models provide a strong baseline for this dataset. Our work compliments the current efforts in NLI by presenting thorough experiments for the specialized and knowledge intensive field of medicine. We also demonstrated that a simple use of domain-specific word embeddings provides a performance boost. Finally, we also presented a method for integrating domain ontologies into the training regime of models. We hope the released code and dataset with clear benchmarks help advance research in clinical NLP and the NLI task. 

\section*{Acknowledgments}
This work would not have been possible without Adam Coy, Andrew Colucci, Chanida Thammachart, and Hassan Ahmad -- the clinicians in our team who helped us in creating the dataset. We are grateful to Vandana Mukherjee and Tanveer Syeda-Mahmood for supporting the project. We would also like to thank Anna Rumshisky and Anna Rogers for their help in this work. Most importantly, we would like to thank Leo Anthony Celi and Alistair Johnson from the MIMIC team for helping us in making MedNLI publicly available.




\bibliography{main}
\bibliographystyle{acl_natbib_nourl}

\end{document}